\documentclass[runningheads]{llncs}
\usepackage[T1]{fontenc}
\usepackage{graphicx}
\usepackage{adjustbox}
\usepackage{booktabs}
\usepackage{tabularx}
\usepackage{multirow}
\usepackage{pifont}
\usepackage{subcaption, soul, xcolor}
\usepackage{hyperref}
\usepackage{amssymb}

\usepackage{acronym}
\newacro{gan}[GAN]{generative adversarial network}
\newacro{vae}[VAE]{variational autoencoder}
\newacro{cnn}[CNN]{convolutional neural network}
\newacro{dire}[DIRE]{diffusion reconstruction error}
\newacro{mlp}[MLP]{multi-layer perceptron}
\newacro{clip}[CLIP]{contrastive language image pre-training}
\newacro{vlm}[VLM]{vision-language model}
\newacro{vit}[ViT]{vision transformer}
\newacro{nlp}[NLP]{natural language processing}
\newacro{cv}[CV]{computer vision}
\newacro{blip}[BLIP]{bootstrapping language image pre-training}
\newacro{vqa}[VQA]{visual question answering}
\newacro{lora}[LORA]{low-rank adaptation}
\newacro{peft}[PEFT]{parameter-efficient fine-tuning}
\newacro{gpt}[GPT]{generative pre-trained transformer}
\newacro{q-former}[Q-Former]{querying transformer}
\newacro{sedid}[SeDID]{stepwise error for diffusion-generated image detection}
\newacro{sd}[SD]{stable diffusion}
\newacro{lsun}[LSUN]{large-scale scene understanding}
\newacro{fc}[FC]{fully-connected}
\newacro{ml}[ML]{machine learning}
\newacro{deit}[DeiT]{data-efficient image transformers}
\newacro{ldm}[LDM]{latent diffusion model}
\newacro{adm}[ADM]{ablated diffusion model}
\newacro{ddpm}[DDPM]{denoising diffusion probabilistic models}
\newacro{iddpm}[IDDPM]{improved denoising diffusion probabilistic models}
\newacro{pndm}[PNDM]{pseudo numerical methods for diffusion models on manifolds}
\newacro{srm}[SRM]{spatial rich model}
\newacro{lasted}[LASTED]{language-guided synthesis detection}
\newacro{rf}[RF]{random forest}
\newacro{dm}[DM]{diffusion model}
\newacro{ddim}[DDIM]{denoising diffusion implicit models}
\newacro{multilid}[multiLID]{multi local intrinsic dimensionality}
\newacro{ifdl}[IFDL]{image forgery detection and localization} 
\newacro{svm}[SVM]{support vector machine} 
\newacro{ai}[AI]{artificial intelligence} 

\newacro{amsff}[AMSFF]{attention-based multi-scale feature fusion} 
\newacro{psm}[PSM]{patch selection module}
\newacro{llm}[LLM]{large language model}
\newacro{coop}[CoOp]{context optimization}
\newacro{fidavl}[FIDAVL]{Fake Image Detect and Attribution using a Vision-Language model}
\newacro{vqa}[VQA]{visual question answering}

\begin{document}
\title{FIDAVL: Fake Image Detection and Attribution using Vision-Language Model}
%
\author{Mamadou Keita\inst{1} \and
Wassim Hamidouche\inst{2} \and
Hessen Bougueffa Eutamene\inst{1} \and 
Abdelmalik Taleb-Ahmed\inst{1}\and
Abdenour Hadid \inst{3}}
\authorrunning{M. Keita {\it et al.}}
%
\institute{Laboratory of IEMN, Univ. Polytechnique
Hauts-de-France, Valenciennes, France \and
Univ. Rennes, INSA Rennes, CNRS, IETR - UMR, Rennes, 6164, France \and
Sorbonne Center for Artificial Intelligence, Sorbonne University Abu Dhabi, UAE}
%
\maketitle              
\begin{abstract}
We introduce \acs{fidavl}: Fake Image Detection and Attribution using a Vision-Language Model. \acs{fidavl} is a novel and efficient multitask approach inspired by the synergies between vision and language processing. Leveraging the benefits of zero-shot learning, \acs{fidavl} exploits the complementarity between vision and language along with soft prompt-tuning strategy to detect fake images and accurately attribute them to their originating source models. We conducted extensive experiments on a comprehensive dataset comprising synthetic images generated by various state-of-the-art models. Our results demonstrate that \acs{fidavl} achieves an encouraging average detection accuracy of 95.42\% and F1-score of 95.47\% while also obtaining noteworthy performance metrics, with an average F1-score of 92.64\% and ROUGE-L score of 96.50\% for attributing synthetic images to their respective source generation models. The source code of this work will be publicly released at \url{https://github.com/Mamadou-Keita/FIDAVL}. 

\end{abstract}
\vspace{-2.0mm}
\section{Introduction}
\label{sec:intro} 
\vspace{-2.0mm}

Over the past two decades, the landscape of techniques for generating and manipulating photorealistic images has undergone rapid evolution. This evolution has ushered in an era where visual content can be easily created and manipulated, leaving behind minimal perceptual traces. Consequently, there is a growing apprehension that we are on the brink of a world where distinguishing real images from computer generated ones will become increasingly challenging. Recent advancements in generative models have further propelled the quality and realism of synthesized images, enabling their application in conditional scenarios for contextual manipulation and broadening the scope of media synthesis. However, amidst these advancements, a prevailing concern persists regarding the potential repercussions of these technologies when wielded maliciously. This apprehension has garnered significant public attention due to its disruptive implications for visual security, legal frameworks, political landscapes, and societal norms~\cite{liz2024generation}. Therefore, it is paramount to delve into the development of effective visual forensic techniques capable of mitigating the threats posed by these evolving generative patterns.

To tackle the challenges posed by generative models, several solutions have emerged in the literature. Existing methodologies predominantly revolve around binary detection strategies (real vs. \acs{ai}-generated)~\cite{cozzolino2023raising,xi2023ai} aimed at discerning synthetic images from authentic ones. However, the task of attributing a generated image to its originating source remains relatively unexplored and inherently complex. With the current level of realism achieved by modern generative models, traditional methods reliant on human inspection for attribution have become impractical. While identifying whether an image was generated by a specific model may seem straightforward, it presents nuanced challenges. A simplistic approach involves training a classifier on a dataset comprising both real and generated images produced by the model in question. However, such an approach is susceptible to dataset bias ~\cite{torralba2011unbiased} and may struggle to generalize effectively when applied to new data. Furthermore, detectors tailored to specific generative models risk obsolescence as generation techniques evolve and the model they were trained on becomes outdated.

Pre-trained large vision-language models have recently emerged as a promising solution for a multitude of natural language processing and computer vision tasks. These models undergo training on vast image-text datasets sourced from the Internet and exhibit proficiency as zero-shot and few-shot learners for downstream tasks, particularly in applications like image classification~\cite{zhang2022tip}, detection~\cite{ming2022delving}, and segmentation~\cite{zhou2023zegclip}. Moreover, there has been a recent surge in leveraging these models for the detection of synthetic images~\cite{cozzolino2023raising,chang2023antifakeprompt,keita2024bi}. 

In the current state-of-art, the detection and attribution of synthetic images often face significant challenges. One of the main difficulties lies in the fact that these tasks are typically handled separately, which can lead to ineffective and less robust solutions. Multi-level or cascade architectures are commonly proposed to address these tasks, but they introduce complexity and can be difficult to generalize across different types of synthetic images. The separation of detection and attribution tasks overlooks the potential synergies that could be leveraged by treating them as related tasks. Additionally, the generalization capabilities of existing models are often limited, which hampers their effectiveness in handling diverse and evolving state-of-the-art image generation techniques.

To address these challenges, we introduce \acs{fidavl}, a novel and efficient multitask method that combines synthetic image detection and attribution within a unified framework. Leveraging a vision-language approach, \acs{fidavl} harnesses synergies between vision and language models along with a soft adaptation strategy. This integration enables precise detection and accurate attribution of generated images to their original source models, capitalizing on shared features between the two tasks. Our approach benefits from the generalization capabilities of \acp{vlm}, which represents a significant advancement over traditional methods. By treating synthetic image detection and attribution as related tasks within a single-step process, \acs{fidavl} overcomes the limitations of multi-level or cascaded architectures. Extensive experiments conducted on a large-scale dataset including synthetic images generated by various state-of-the-art models demonstrate the high accuracy and robustness of \acs{fidavl}. This approach not only simplifies the process of detection and attribution but also enhances its reliability and scalability. To the best of our knowledge, this study pioneers the utilization of vision-language models for synthetic image attribution and detection in a unified framework.

Our contributions to this paper can be summarized as follows:
\vspace{-2.0mm}
\begin{itemize}
\item[\ding{112}] We introduce \acs{fidavl}, a novel single-step approach for synthetic image detection and attribution. Leveraging the complementarity between vision and language, \acs{fidavl} effectively detects and attributes synthetic images to their respective source generation models.
\item[\ding{112}] We adopt a soft prompt-tuning technique to refine the query of \acs{fidavl} for optimal effectiveness. 
\end{itemize}

Through extensive evaluation on a large-scale dataset, our proposed approach demonstrates competitive performance, underscoring its effectiveness in  synthetic image detection and attribution.  \acs{fidavl} achieves an average accuracy (ACC) exceeding 95\% in the synthetic image detection task, and yielding an average ROUGE-L score of 96.50\% and an F1-score of 92.64\% in the synthetic image attribution task.

The remainder of this paper is organized as follows. Section~\ref{sec:relatedWork} provides a brief review of the background and related work. Section~\ref{sec:Papproach} describes the proposed \acs{fidavl} approach for the attribution and detection of synthetic images. Then, the performance of the proposed approach is assessed and analysed in Section~\ref{sec:resultsAnalysis}. Finally, Section~\ref{sec:conclusion} concludes the paper.

\vspace{-2.0mm}
\section{Background and Related Work}
\label{sec:relatedWork}
\vspace{-2.0mm}

In this section, we delve into generative models, examine advanced deepfake detection and attribution techniques, and offer insights into vision-language models and prompt tuning.

\subsection{Generative Models} 
Generative models have emerged as powerful tools for synthesizing realistic data across various modalities, including images, text, videos, and intricate structures. These models, often harnessed through neural networks, adeptly learn to capture and replicate the underlying patterns and distributions inherent in the training data~\cite{esser2021taming}. Within the domain of deep generative models, a prominent category is  \ac{gan}~\cite{goodfellow2014generative}. More recently, diffusion models~\cite{song2020denoising} have gained traction as a de-facto method for image generation. The extension of such models to text-to-image synthesis~\cite{rombach2022high,nichol2021glide} has ushered in a wave of models characterized by remarkable quality and diversity, exemplified by models like Imagen~\cite{saharia2022photorealistic} and DALL-E-2~\cite{ramesh2022hierarchical}. However, the proliferation of deep generative models in image synthesis has also given rise to challenges pertaining to synthetic image detection and attribution.

\subsection{Synthetic Image Detection and Attribution}
Recent strides in generative models, particularly diffusion-based architectures and cutting-edge \acs{gan} models, present challenges to existing detection methodologies. Research highlighted in~\cite{corvi2023detection,ricker2022towards} underscores the struggle of current detectors to adapt to these innovative models, underscoring the need for more effective detection techniques. Consequently, a spectrum of novel approaches has emerged in response. Coccomini {\it et al.}~\cite{coccomini2023detecting} experiment with \acp{mlp} and conventional \acp{cnn}, probing their efficacy in this domain. Conversely, Wang {\it et al.}\cite{wang2023dire} introduce \acs{dire}, a method tailored for diffusion-generated images, which prioritizes the analysis of reconstruction errors. Leveraging diffusion patterns, \acs{sedid}~\cite{ma2023exposing} achieves accurate detection, with a focus on reverse and denoising computation errors. Amoroso {\it et al.}~\cite{amoroso2023parents} explore semantic-style disentanglement to bolster stylistic detection, while Xi {\it et al.}~\cite{xi2023ai} propose a dual-stream network that emphasizes texture for \ac{ai}-generated image detection. Wu {\it et al.}~\cite{wu2023generalizable} advocate for \ac{lasted}, treating detection as an identification problem and leveraging language-guided contrastive learning. Ju {\it et al.}~\cite{ju2023glff} propose a feature fusion mechanism, combining ResNet50 and attention-based modules, for global and local feature fusion in \ac{ai}-synthesized image detection. Sinitsa {\it et al.}~\cite{sinitsa2023deep} introduce a rule-based method harnessing \acp{cnn} to extract distinctive features, achieving high accuracy even with limited generative image data. In a departure from traditional approaches, Chang {\it et al.}~\cite{chang2023antifakeprompt} draw from \acp{vlm}, framing deepfake detection as a visual question-answering task. Finally, Cozzolino {\it et al.}\cite{cozzolino2023raising} propose a lightweight strategy based on \ac{clip} features and linear \ac{svm}, showcasing an alternative avenue for effective detection in this rapidly evolving landscape.

Attributing deepfake content to its source constitutes a crucial aspect in the realm of detection and prevention. Unlike conventional binary detection, attribution introduces a multi-class dimension, facilitating the identification of the specific generative model responsible for the content. Recent studies have shed light on the importance of enhancing attribution techniques. He {\it et al.}\cite{he2023mgtbench} extended detectors to explore textual attribution, revealing areas ripe for improvement in this domain. In the realm of generative visual data, attribution methodologies tailored for \acp{gan} have emerged. Bui {\it et al.}~\cite{bui2022repmix} introduced a \acs{gan}-fingerprinting technique, which notably enhances source attribution in a closed-set scenario. Recent advancements have also focused on diffusion models (\acp{dm}). Sha {\it et al.}~\cite{sha2022fake} utilized ResNet for detecting and attributing synthetic images to their respective generators, while Guarnera {\it et al.}~\cite{guarnera2023level} proposed a multi-level approach for synthetic image detection and attribution. Lorenz {\it et al.}~\cite{lorenz2023detecting} introduced \acs{multilid}, a method tailored for diffusion-generated image detection and attribution, leveraging intrinsic dimensionality for enhanced accuracy. Moreover, Wang {\it et al.}~\cite{wang2023evaluating} addressed the attribution of generative data to their training data counterparts, necessitating the identification of significant contributors within the training set. 

\vspace{-2.0mm}
\subsection{Vision Language Models}
\label{sec:vlm}
\vspace{-1.0mm}

Recent advancements in \acp{vlm} have addressed limitations inherent in earlier models, particularly in terms of task specificity and dataset constraints. Noteworthy models such as \ac{clip}, trained on an extensive dataset comprising 400 million image-caption pairs, exemplify this progress by featuring both image and text encoders, thereby facilitating versatile image classification tasks. Leading the charge in this domain are pioneering models such as LLaVA~\cite{liu2023visual}, BLIP2~\cite{li2023blip}, InstructBLIP~\cite{dai2023instructblip}, and Flamingo~\cite{alayrac2022flamingo}, which represent the vanguard of \acp{vlm} innovation. LLaVA, an open-source endeavor, seamlessly integrates vision and language understanding within a vast multimodal framework. BLIP2, on the other hand, achieves state-of-the-art performance through the integration of pre-trained image encoders and language models. Building upon BLIP2, InstructBLIP refines its architecture further, specifically tailoring it for visual instruction tuning. Notably, Flamingo, a family of \acp{vlm}, stands out for its adeptness in handling interleaved visual and textual data, thereby making significant strides in adapting to downstream tasks and expanding zero-shot capabilities. These advancements mark a significant leap forward in the realm of \acp{vlm}, showcasing their potential to revolutionize various domains reliant on multimodal understanding and processing.

\vspace{-2.0mm}
\subsection{Prompt Tuning for Vision Language Models}
\label{sec:PTuning}
\acp{vlm} excel in learning from multimodal data, yet encounter challenges when tasked with adapting to specific downstream vision-related objectives. Groundbreaking research by~\cite{zhou2022learning} introduced \ac{coop} to augment the efficiency of \acs{clip} in image classification tasks. Diverging from conventional prompt templates, \ac{coop} learns prompt embeddings with minimal reliance on downstream dataset samples. Prompt tuning manifests in two primary forms: hard and soft. Hard prompt tuning, as proposed in~\cite{zou2023universal}, involves adjusting non-differentiable tokens to align with user-defined criteria, albeit encountering difficulties in achieving discrete improvements. Conversely, soft prompt tuning, showcased by~\cite{lester2021power}, optimizes a trainable tensor through back-propagation, thereby enhancing modeling performance. In a notable application,~\cite{chen2023instructzero} employed subtle prompt optimization techniques to enhance instruction generation in a black-box \ac{ml} model. These endeavors underscore the importance of nuanced prompt tuning methodologies in enhancing the adaptability and performance of vision-language models across various downstream tasks.

\vspace{-2.0mm}
\section{Proposed Synthetic Image Detection and Localization}
\label{sec:Papproach}
\vspace{-2.0mm}
\subsection{Problem Formulation}
\label{sec:problemFormulation}

\begin{figure*}[!ht]
    \centering
    \includegraphics[width=1\linewidth]{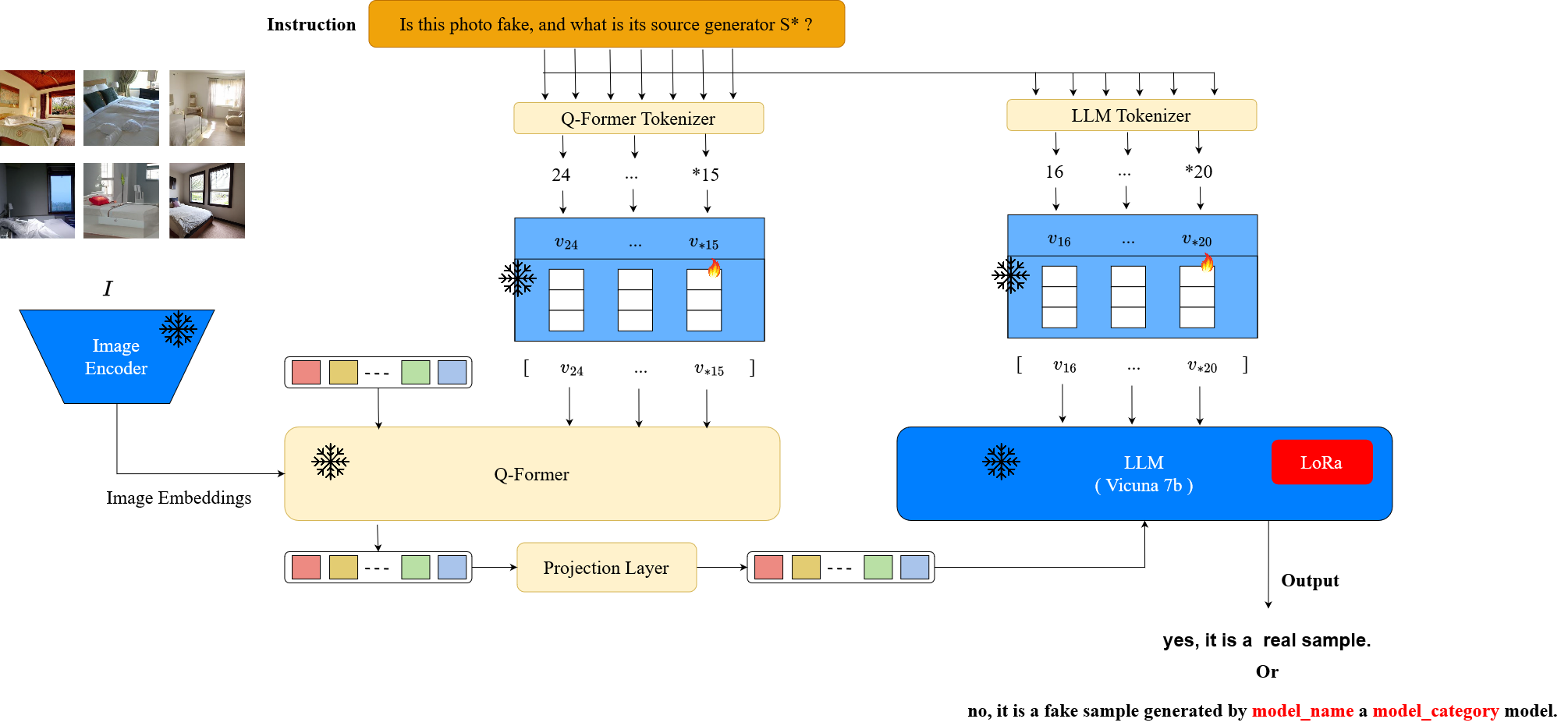}
    \caption{Architecture of the proposed synthetic image detection and localization.}
    \label{fig:architecture}
\end{figure*}

To harness the capabilities of a vision-language model, such as InstructBLIP, we have embraced a framework known as \ac{vqa}, which we refer to as \acs{fidavl}. \acs{fidavl} is meticulously crafted to respond to inquiries regarding a given image. The input comprises two crucial components: a query image, denoted as $I$, which serves as the focal point of our scrutiny, and a composite question, denoted as $q$, which guides \acs{fidavl} in its analysis of the query image. Subsequently, the image is classified as either real or fake; if fake, it is then attributed to its source. The question $q$ can take on various forms, ranging from predefined inquiries like "Is this photo fake, and what is its source generator?" to customizable questions incorporating a pseudo-word $S^*$. This adaptability empowers us to tailor our questioning strategy to the specific requirements of our investigation.

The output of \acs{fidavl} comprises a set of response texts, denoted as $\hat{y}$. While $\hat{y}$ theoretically encompasses any text, we impose specific constraints to uphold consistency and clarity in our responses. If the query image is determined to be real, the response is articulated as \textbf{"No, it is a real sample."}. Conversely, if it is deemed fake, the response adheres to the template \textbf{"Yes, it is a fake sample generated by \textit{model\_name}, a \textit{model\_category} model."}. Here, \textit{model\_name} signifies the name of the generating model, which could belong to the set { progan, diff-projectedgan, stylegan, ldm, glide, Stable diffusion}, while \textit{model\_category} denotes the category of the generating model, which could be {diffusion or gan}. This response structure aligns with our ground truth for synthetic image detection and attribution. Finally, to evaluate the efficacy of \acs{fidavl}, we gauge the accuracy of both the detection and attribution tasks. This quantitative assessment offers insights into our model's proficiency in accurately identifying and attributing synthetic images.

Mathematically, the formulation of the single-step synthetic image detection and attribution task is as follows:

\begin{equation}
    \hat{y} = \mathcal{M}_{\theta}(I,q).
\end{equation}

where $\mathcal{M}$ is an \acs{vlm} with parameters $\theta$, which takes an image $I$ and a question $q$ as input and generates an answer $\hat{y}$. 

\subsection{Soft Prompt Tuning}
\label{sec:Softfinetuning}
\vspace{-1.0mm}
Our investigation harnesses soft prompt tuning within InstructBLIP, following the outlined procedure. In InstructBLIP, the prompt serves as input to two pivotal components: Q-Former and \ac{llm}. Initially, the prompt undergoes tokenization and embedding before being concurrently fed into both Q-Former and the \ac{llm}, as illustrated in Fig.~\ref{fig:architecture}. To facilitate prompt tuning, we introduce a pseudo-word $S^*$ into the prompt, which acts as the target for tuning. Specifically, we adopt the question pattern "Is this photo fake, and what is its source generator?", appending the pseudo-word to the end of the prompt. This modification yields the following adjusted prompt $q^*$: "Is this photo fake, and what is its source generator $S^*$?". For real images, we assign the output label $y$ as "No, it is a real sample." Conversely, for fake images, the label $y$ is set as "Yes, it is a fake sample generated by \textit{model\_name}, a \textit{model\_category} model." This labeling scheme facilitates soft prompt tuning.

We then proceed to freeze all model modules except the word embedding $v^*$ corresponding to the pseudo-word $S^*$, which is randomly initialized. Subsequently, we optimize the word embedding $v^*$ of the pseudo-word across a triplet training set \{$I, q^*, y$\} using the language modeling loss. Our aim is to align the output of the \ac{vlm}, denoted as $\hat{y}$, with the label $y$. Our optimization objective can therefore be defined as :

\begin{equation}
    f_{S^*} = \arg\min_{S^*} \mathbb{E}_{(I, y)} \left[ L(M(I, q^*), y) \right]
\end{equation}

where $L$ is the language modeling loss function (cross-entropy loss).

\vspace{-2.0mm}
\section{Experimental Results}
\label{sec:resultsAnalysis}
\vspace{-2.0mm}

{\bf Dataset.} The dataset utilized in this study is a meticulously curated collection of images comprising two primary components: real images sourced from the \ac{lsun} bedroom dataset and synthetic data generated by three distinct \ac{gan} engines (ProGAN, StyleGAN, Diff-ProjectedGAN), as well as three text-to-image \ac{dm} models (LDM, Glide, Stable diffusion v1.4). For each considered \ac{gan}, 20,000 images were generated for training and an additional 10,000 for testing, resulting in a total of 90,000 synthetic images. Similarly, each \acs{dm} architecture generated an equivalent number of images for both training and testing, leveraging the prompt "A photo of a bedroom", thus yielding another 90,000 images. Consequently, the cumulative synthetic dataset comprises 180,000 images. In addition to synthetic data, the dataset incorporates 130,000 real images. Notably, the real images designated for testing remain consistent across all testing subsets. \\\\ 
\textbf{Implementation Details.} We use the GitHub repository of~\cite{chang2023antifakeprompt} based on LAVIS library for implementation, training, and evaluation. To prevent out-of-memory issues on small GPU, we employ Vicuna-7B as \acs{llm}. For prompt tuning, we initialize the model with an instruction-tuned checkpoint from LAVIS, exclusively fine-tuning the word embeddings of the pseudo-word while freezing the rest of the model. The model is prompt-tuned with a maximum of $5$ epochs, employing the AdamW optimizer with $\beta_1 = 0.9$ and $\beta_2 = 0.999$, batch size $16$, and a weight decay of $0.05$. The initial learning rate is set to $10^{-8}$, and apply cosine decay with a minimum learning rate of $0$. The code is executed on an NVIDIA RTX A4500 GPU with 16 GB and an Intel(R) i9-12950HX CPU with Windows 11 Pro. In terms of image processing, all the images are resized to 224 pixels on the shorter side, maintaining the original aspect ratio. In training, random cropping yields a final size of 224$\times$224 pixels, while testing involves center cropping to the same size. \\\\
\textbf{Evaluation Metrics.} In our synthetic image detection and attribution task, we evaluate our \acs{fidavl} model across multiple metrics including accuracy, F1-score. Since we cannot directly compare results from textual data as if it were binary classification, what we can do is calculate overlapping words between predictions and references. In this regard, we use the ROUGE score, which measures the degree of correspondence between the content of the generated sentence and the content of a set of reference sentences. The higher the value of these metrics, the better the performance of the model.

\subsection{Synthetic Image Detection}
In this section, we delve into an extensive analysis of these results, meticulously examining the model's performance across our test set and elucidating the strengths of our detection strategy. Through a comprehensive examination of metrics such as accuracy (ACC) and F1 score, we aim to gain deeper insights into the efficacy with which \acs{fidavl} tackles the task of synthetic image detection.

Table~\ref{tab:detectionTask} showcases the evaluation outcomes concerning the detection capabilities of our proposed method, \acs{fidavl}. Across all test subsets, \acs{fidavl} showcased robust performance, consistently attaining high accuracy and F1 scores. Remarkably, \acs{fidavl} achieved an average accuracy of \textbf{95.42\%} alongside an impressive F1 score of \textbf{95.47\%}, underscoring its effectiveness in precisely distinguishing between synthetic and authentic images.

\begin{table*}[]
\centering
\caption{Synthetic image detection task and comparison to baseline models. We report ACC (\%) / F1-Score (\%). Note that, on average (two last columns), our model yields better performance.}
\label{tab:detectionTask}
{\begin{adjustbox}{max width=\linewidth}
\begin{tabular}{@{}l|c|c|c|c|c|c|cc@{}}
\toprule
\multirow{2}{*}{Method} & \multicolumn{6}{c}{Testing Subset} & \multirow{2}{*}{\begin{tabular}[c]{@{}c@{}}Average\\ (in \%)\end{tabular}} \\ \cmidrule(lr){2-7}
       & LDM$^\star$    & SD v1.4$^\star$  & GLIDE$^\star$   & ProGAN $^\oplus$   & StyleGAN$^\oplus$  & Diff-ProjectedGAN$^\oplus$   &   \\ \cmidrule(r){1-1} \cmidrule(l){8-8} 
ResNet50 & 99.92 / 99.92 & 75.47 / 67.57 & 73.10 / 63.28 & 94.28 / 93.94 & 77.94 / 71.75 & 59.20 / 31.27 & 79.98 / 71.29 \\
Xception & \bf 99.96 / 99.96 & 63.84 / 43.41  & 58.92 / 30.35 & 64.50 / 45.11 & 69.96 / 57.18 & 51.14 / 04.79 & 68.05 / 46.80 \\
\acs{deit} & 99.83 / 99.83 & 96.02 / 95.86 & \bf 98.15 / 98.11 & 93.28 / 92.81 & 95.08 / 94.84 & 77.06 / 70.32 & 93.23 / 91.96 \\  \midrule
FIDAVL   & 90.84 / 90.62 & \bf 96.53 / 96.64 &  96.56 / 96.67 & \bf 96.56 / 96.67 & \bf 95.83 / 95.94 &  \bf 96.20 / 96.31 &  \bf 95.42 / 95.47\\
\bottomrule
\end{tabular}
\end{adjustbox}}
 \begin{flushleft}
    \scriptsize $^\star$ Diffusion-based model.  $^\oplus$ GAN-based model.
\end{flushleft}
\end{table*}

The efficacy of \acs{fidavl} can be attributed to its innovative approach, leveraging the complementary strengths inherent in vision and language modalities. By seamlessly integrating both vision and language models, \acs{fidavl} harnesses the semantic understanding embedded within each modality, enabling it to discern nuanced cues and patterns indicative of synthetic image generation. This underscores the significance of interdisciplinary methodologies in crafting resilient solutions to intricate challenges like synthetic image detection.

\begin{figure*}[!ht]
      \centering
      \begin{subfigure}[b]{0.3\linewidth}
        \includegraphics[width=\linewidth]{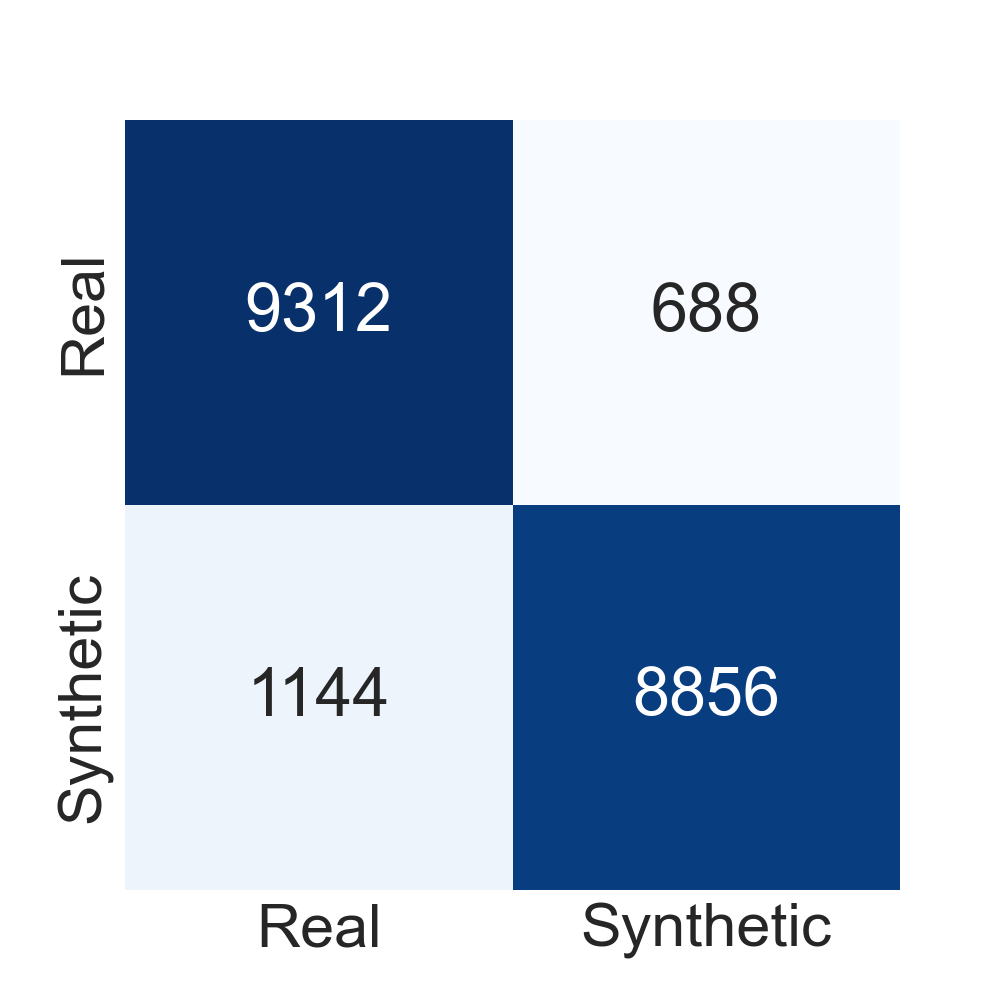}
        \caption*{LDM}
      \end{subfigure}
      \begin{subfigure}[b]{0.3\linewidth}
        \includegraphics[width=\linewidth]{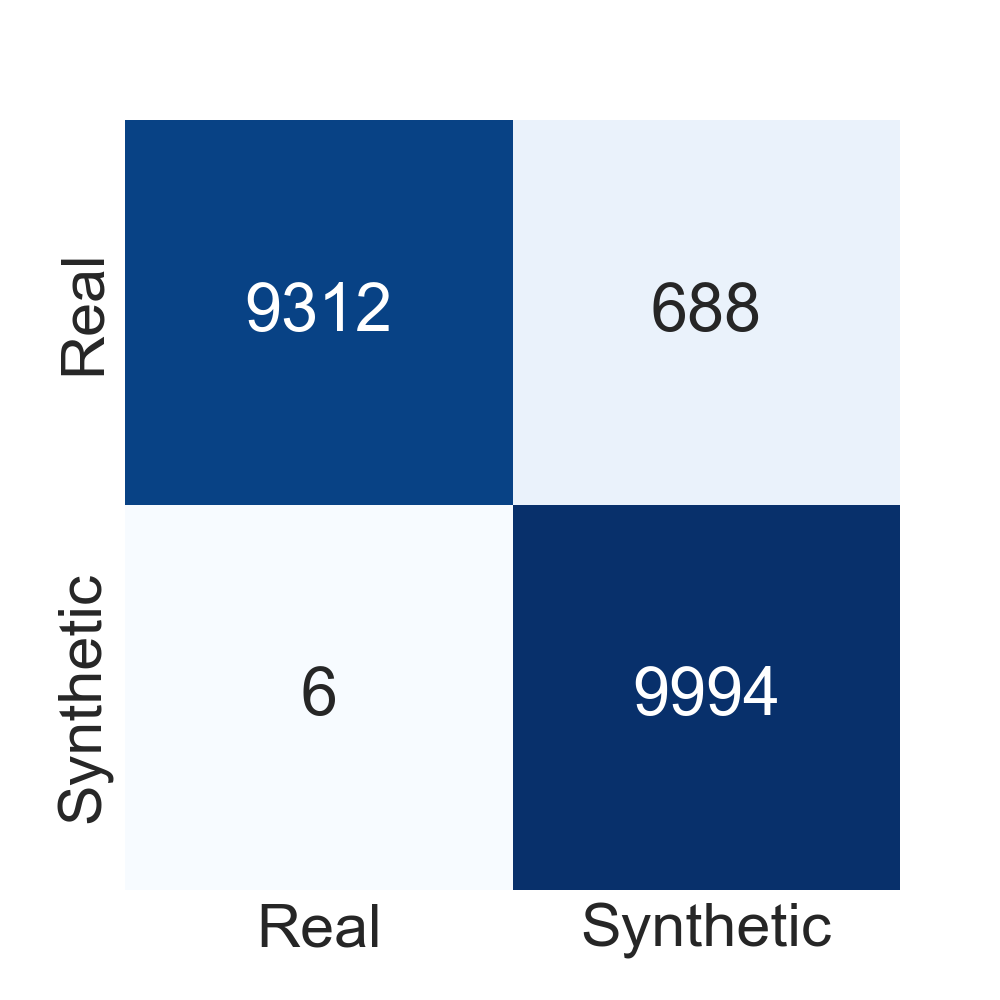}
        \caption*{Stable Diffusion}
      \end{subfigure}
      \begin{subfigure}[b]{0.3\linewidth}
        \includegraphics[width=\linewidth]{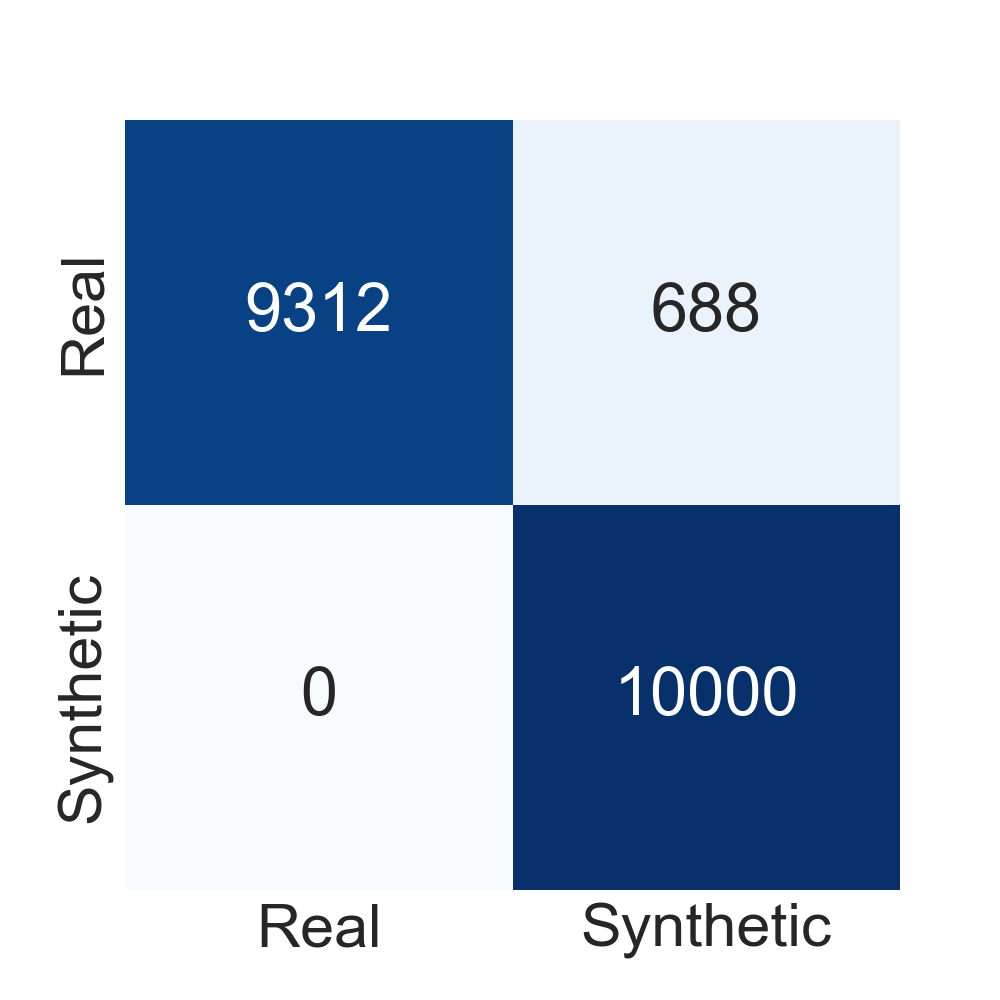}
        \caption*{Glide}
      \end{subfigure}
      \begin{subfigure}[b]{0.3\linewidth}
        \includegraphics[width=\linewidth]{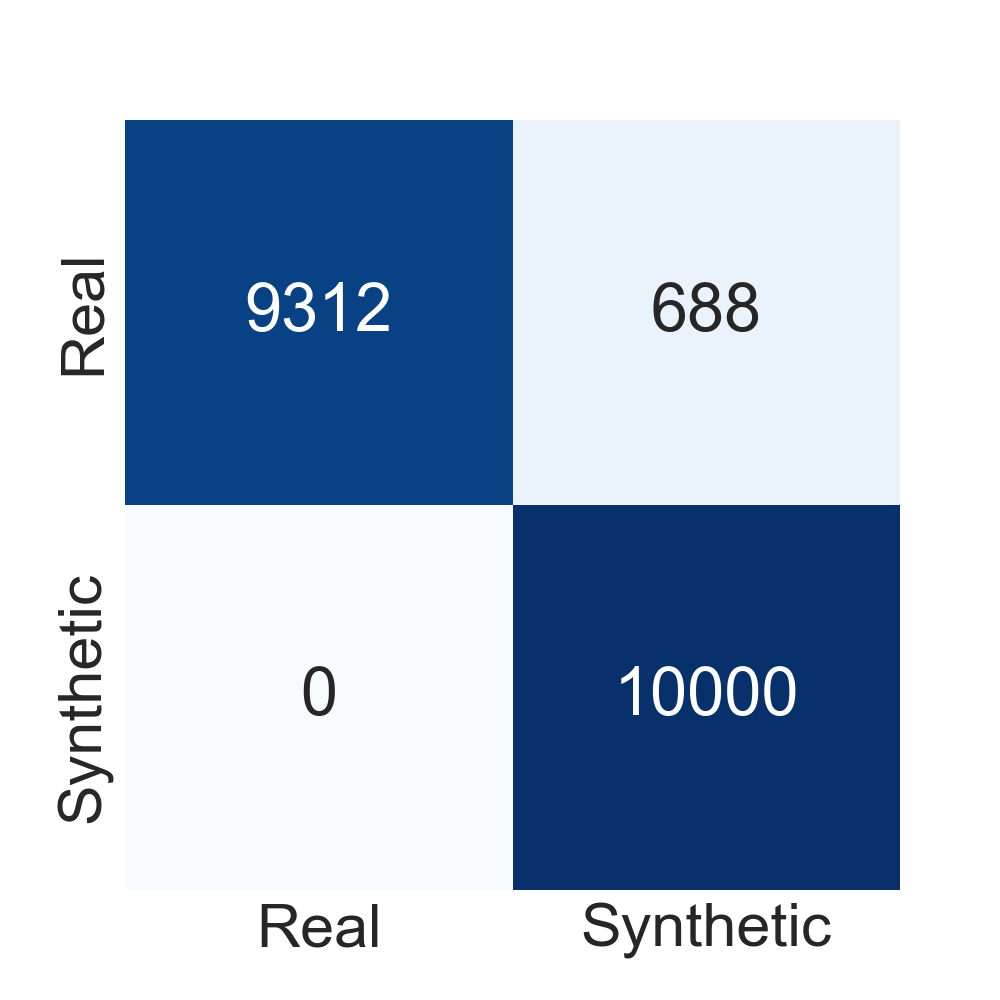}
        \caption*{ProGAN}
      \end{subfigure}
      \begin{subfigure}[b]{0.3\linewidth}
        \includegraphics[width=\linewidth]{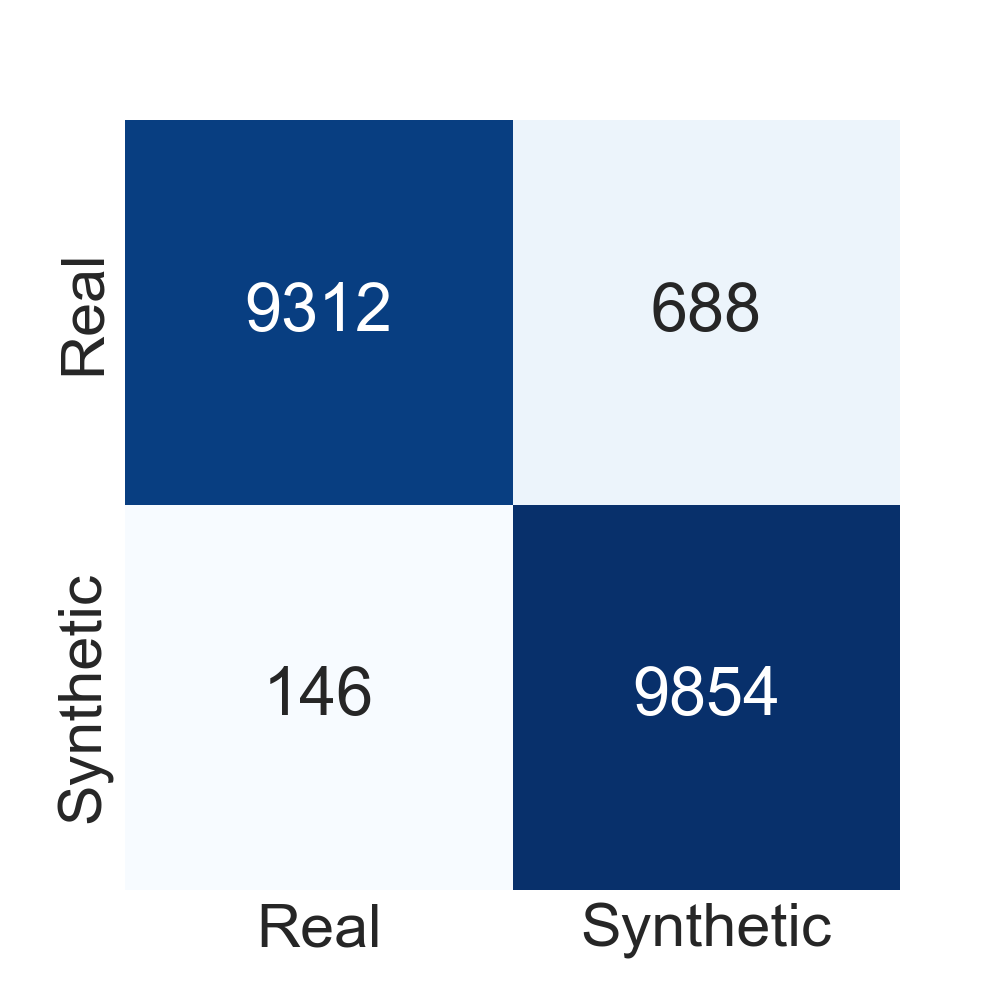}
        \caption*{StyleGAN}
      \end{subfigure}
      \begin{subfigure}[b]{0.3\linewidth}
        \includegraphics[width=\linewidth]{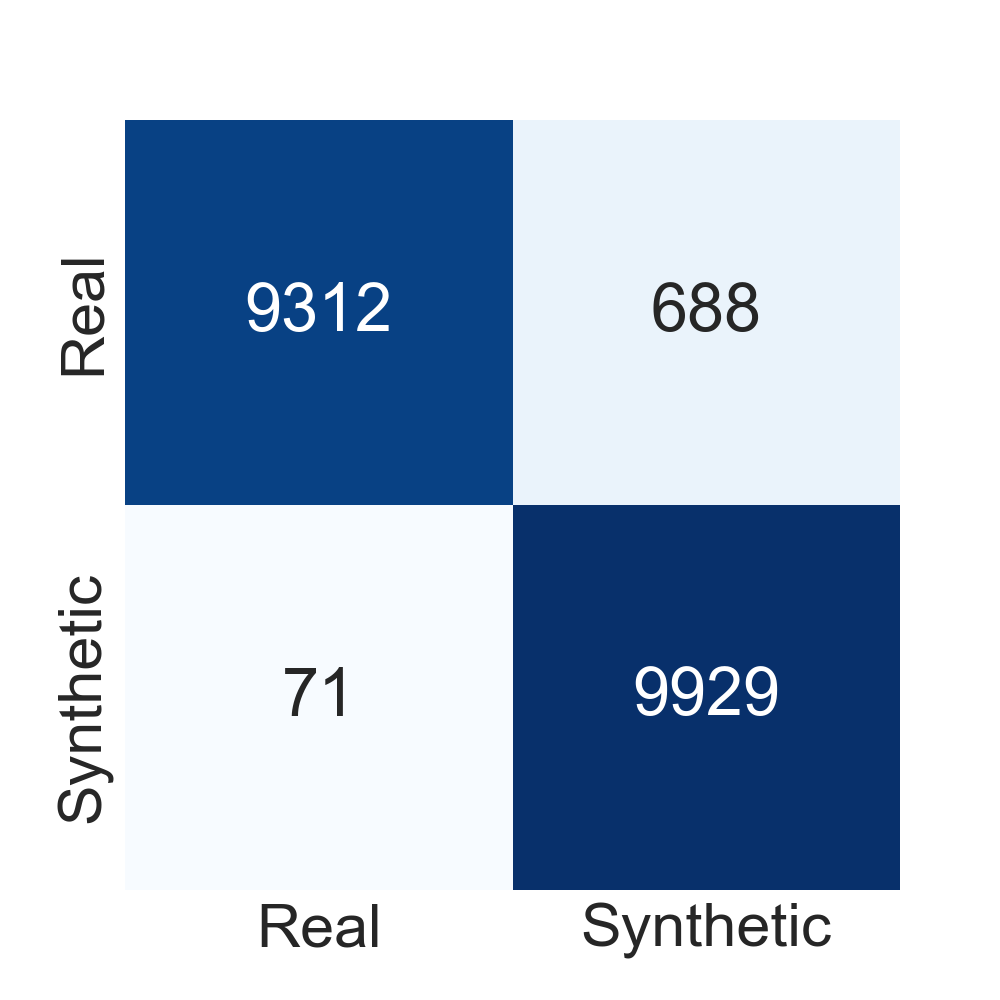}
        \caption*{Diff-ProjectedGAN}
      \end{subfigure}
      \caption{Confusion matrices per testing subset on synthetic image detection task.}
      \label{fig:confusion-matrix}
\end{figure*}

Fig.~\ref{fig:confusion-matrix} provides a comprehensive overview of \acs{fidavl}'s performance in differentiating synthetic image samples from real ones. Each subfigure depicts a confusion matrix corresponding to a specific testing subset, labeled accordingly. Across all subsets, a consistent false negative rate of $688$ is observed, underscoring a shared challenge in accurately detecting synthetic images. Notably, the most promising results are observed in the glide and progan subsets, where all synthetic images were detected. However, \acs{fidavl} encounters challenges in accurately detecting \acs{ldm}-generated images, as evidenced by a significant number of true positives, totaling $1144$. This difficulty can be attributed to the homogeneity of our specific bedroom image dataset, which presents distinct characteristics that may pose challenges for detection algorithms.

\begin{figure*}[!ht]
      \centering
      \begin{subfigure}[b]{0.3\linewidth}
        \includegraphics[width=\linewidth]{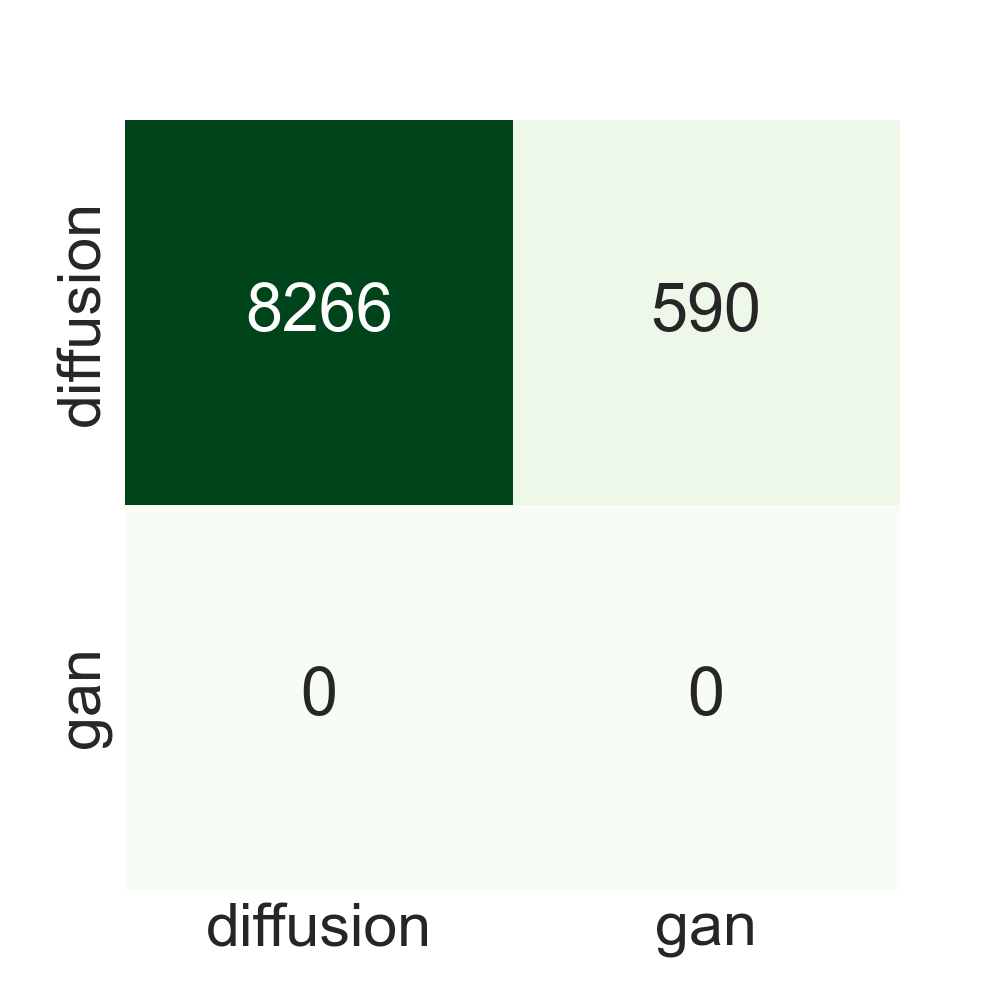}
        \caption*{LDM}
      \end{subfigure}
      \begin{subfigure}[b]{0.3\linewidth}
        \includegraphics[width=\linewidth]{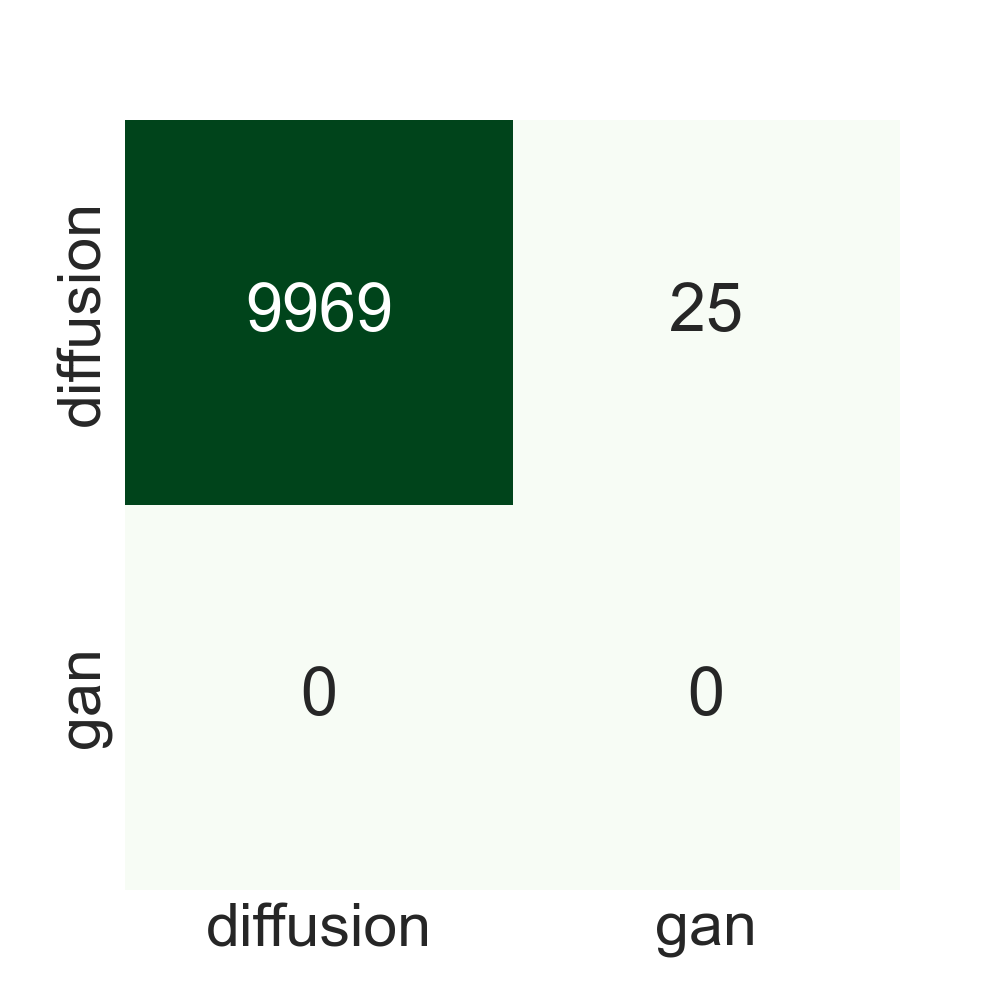}
        \caption*{Stable Diffusion}
      \end{subfigure}
      \begin{subfigure}[b]{0.3\linewidth}
        \includegraphics[width=\linewidth]{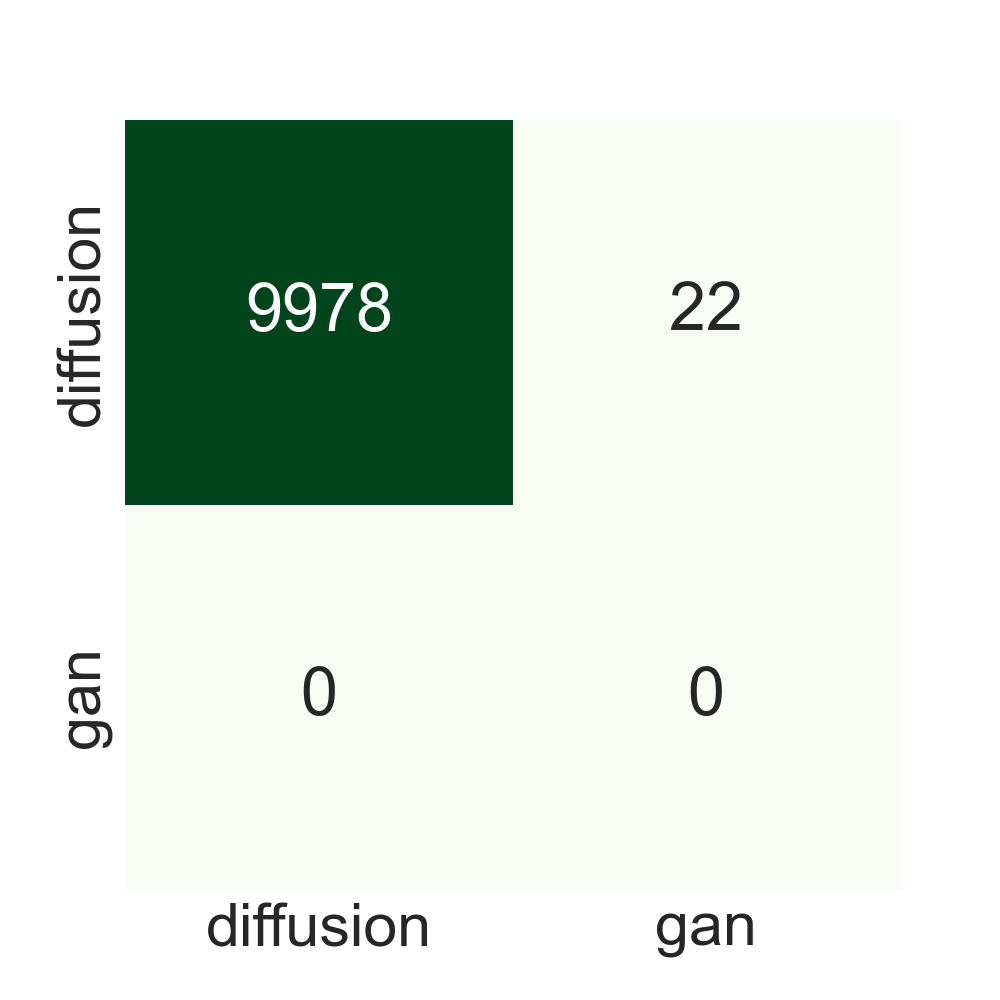}
        \caption*{Glide}
      \end{subfigure}
      \begin{subfigure}[b]{0.3\linewidth}
        \includegraphics[width=\linewidth]{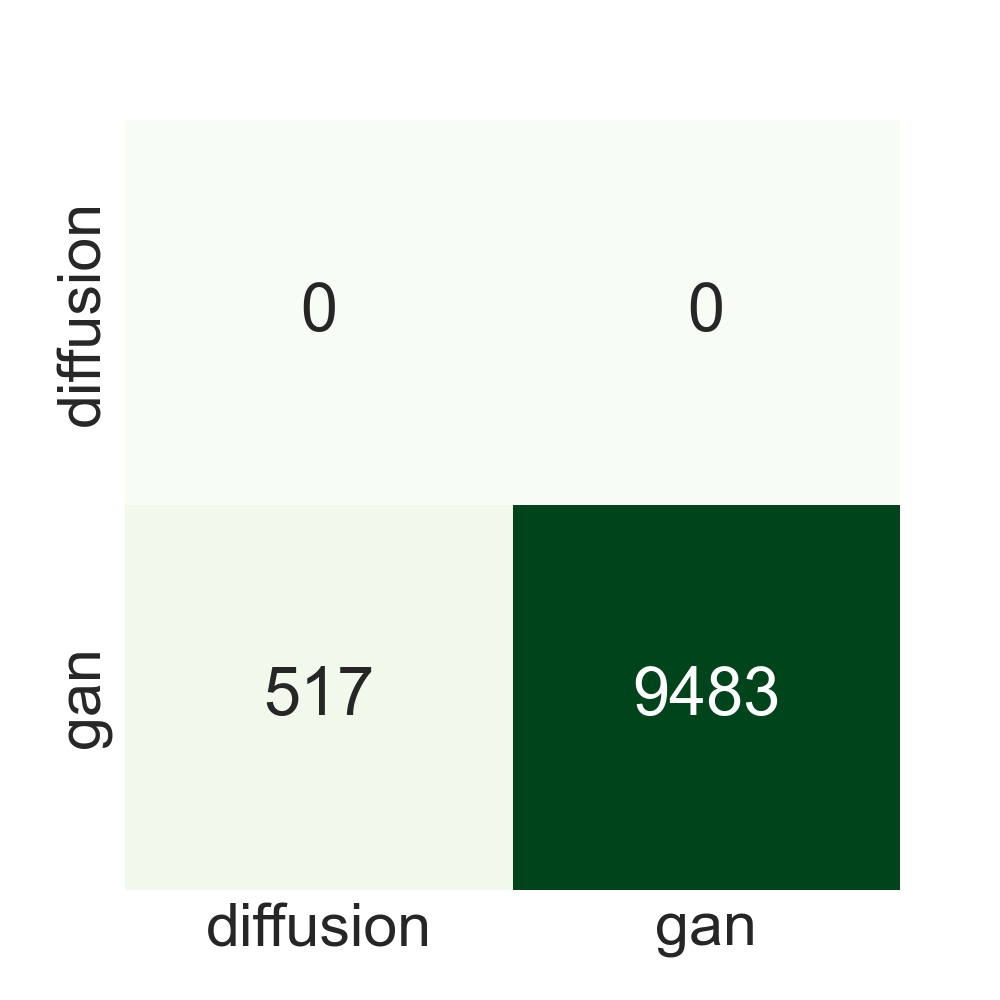}
        \caption*{ProGAN}
      \end{subfigure}
      \begin{subfigure}[b]{0.3\linewidth}
        \includegraphics[width=\linewidth]{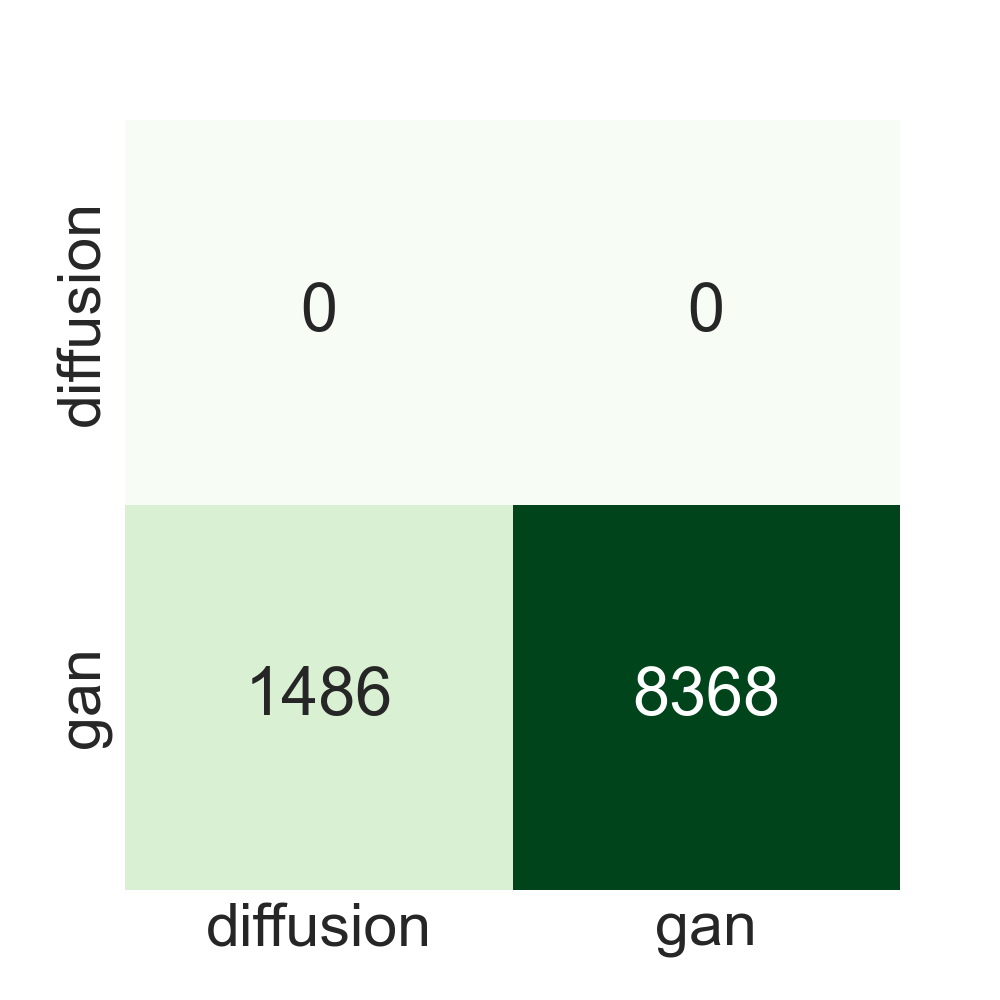}
        \caption*{StyleGAN}
      \end{subfigure}
      \begin{subfigure}[b]{0.3\linewidth}
        \includegraphics[width=\linewidth]{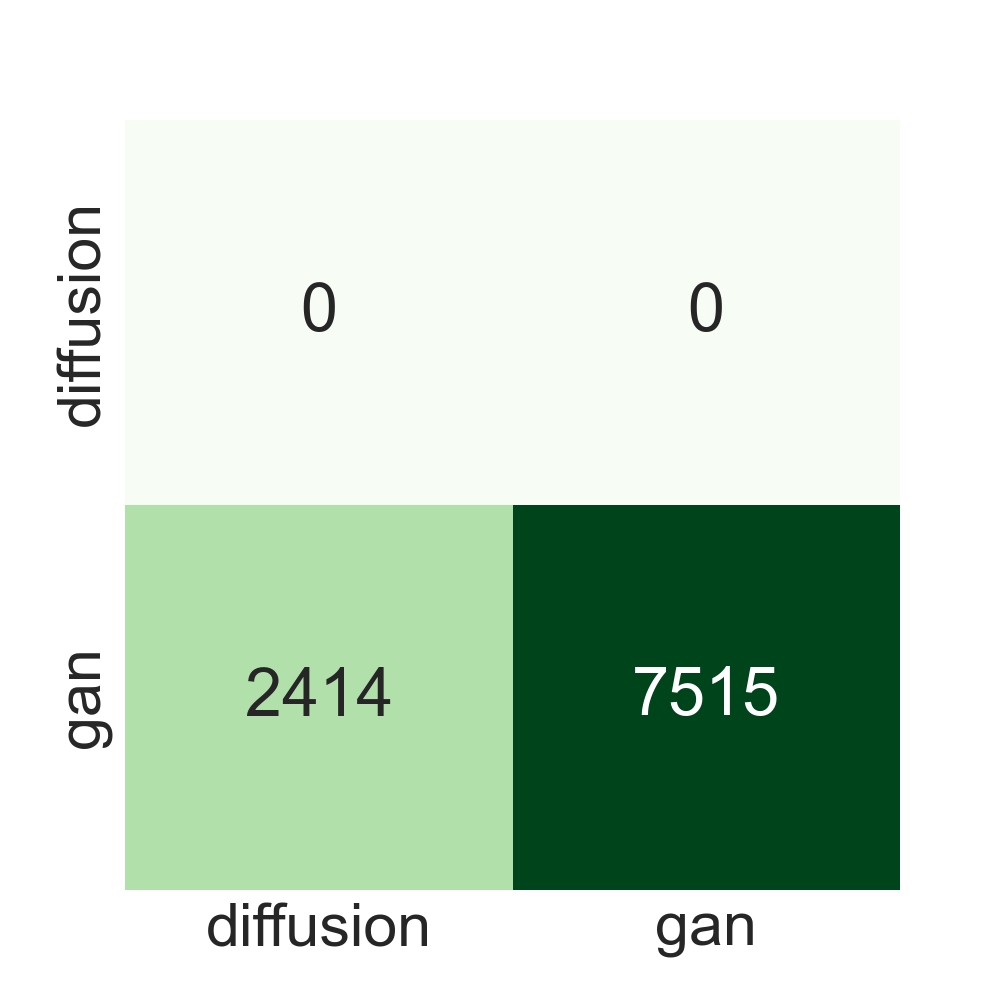}
        \caption*{Diff-ProjectedGAN}
      \end{subfigure}
      \caption{Confusion matrices indicate which synthetic images detected as synthetic are correctly classified according to their generating source model.}
      \label{fig:confusion-matrix-diffusion-gan}
\end{figure*}

Fig.~\ref{fig:confusion-matrix-diffusion-gan} provides an in-depth analysis of the distribution of well-detected synthetic images according to whether they were generated by \acs{gan}-based or diffusion-based models. In Fig.~\ref{fig:confusion-matrix}, we observed from the \acs{ldm} confusion matrix that $8856$ synthetic images were well detected. Furthermore, in Fig.~\ref{fig:confusion-matrix-diffusion-gan}, the \acs{ldm} confusion matrix illustrates the distribution of these images based on their attribution to the respective generator source model type, $8266$ to diffusion and $590$ to \acs{gan}. Fig.~\ref{fig:confusion-matrix-diffusion-gan} shows that although the images have been well classified as synthetic, \acs{fidavl} encounters challenges in accurately attributing these images to their specific source model type, a phenomenon particularly observed with \acs{gan}-based test sets and \acs{ldm}. Moreover, the best performances are obtained on stable diffusion and glide.

\vspace{-5mm}
\subsubsection{Comparative analysis.}
In this subsection, we conduct a comparative analysis of \acs{fidavl} against three baseline models: ResNet50, Xception, and DeiT. To establish our baseline models, we fine-tuned these architectures by replacing their final \acs{fc} layers with a novel \acs{fc} layer containing a single neuron dedicated to distinguishing real images from fake ones. These models were initialized with pre-trained weights obtained from the ImageNet dataset, thereby leveraging the knowledge encoded in their learned representations. We evaluate each model's performance across multiple testing subsets, including LDM, SD v1.4, GLIDE, ProGAN, StyleGAN, and Diff-ProjectedGAN. We present the average performance across these subsets to offer a comprehensive view of the models' effectiveness.

Table \ref{tab:detectionTask} summarized the obtained results from the experiment. ResNet50 performs exceptionally well, particularly in the LDM subset with 99.92\% accuracy and 99.92\% F1 score, and maintains good performance across other subsets with an average accuracy of 79.98\% and F1 score of 71.29\%. Xception shows comparable accuracy in the LDM (99.96\%), but declines considerably in the other subsets, with an average accuracy of 68.05\% and an F1 score of 46.80\%. DeiT demonstrates strong performance, especially in the SD v1.4 (96.02\% accuracy and 95.86\% F1 score) and GLIDE (98.15\% accuracy and 98.11\% F1 score) subsets, with an average accuracy of 93.23\% and an F1 score of 91.96\%. In contrast, \acs{fidavl} exhibits outstanding performance across all subsets, with an average accuracy of 95.42\% and an F1 score of 95.47\%. In particular, \acs{fidavl} excels in SD v1.4, ProGAN, StyleGAN, and Diff-ProjectedGAN subsets, showcasing its robustness and competitiveness compared to the baseline models.

To summarize, our approach shows competitive performance, albeit with lower scores in testing subsets such as LDM and GLIDE. Notably, \acs{fidavl} reaches around 90.84\% on LDM and maintains scores above 95\% on other subsets. \acs{fidavl} adopts a multitask learning approach, which not only involves image detection (distinguishing real from fake) but also includes an attribution task aimed at identifying the model responsible for generating a given image. This dual-focus training introduces additional complexity and objectives to the model's training regimen, which can likely influence its performance dynamics as it must balance learning across multiple objectives.

\vspace{-4mm}
\subsubsection{Generalization to unseen generative models.}
In this subsection, we evaluate \acs{fidavl} generalization capabilities on multiple unseen synthetic image detection subsets, including ADM, DDPM, IDDPM, PNDM, Diff-StyleGAN2, and ProjectedGAN. Each subset represents distinct characteristics and challenges within the detection task, enabling a comprehensive assessment of \acs{fidavl}'s generalization capabilities.

\begin{table*}[]
\caption{Generalization results  on  synthetic images generated by unseen generation models. We report ACC (\%) / F1-Score (\%).}
\label{tab:detectionTaskGeneralization}
\begin{adjustbox}{max width=\linewidth}
\begin{tabular}{@{}l|c|c|c|c|c|c|cc@{}}
\toprule
\multirow{2}{*}{Method} &
  \multicolumn{6}{c}{Testing Subsets} &
  \multirow{2}{*}{\begin{tabular}[c]{@{}c@{}}Average\\ (in \%)\end{tabular}} \\ \cmidrule(lr){2-7}
 &
   ADM$^\star$ &
  DDPM$^\star$ &
  IDDPM$^\star$ &
  PNDM$^\star$ &
  Diff-StyleGAN2$^\oplus$ &
  ProjectedGAN$^\oplus$ &
    \\  \cmidrule(r){1-1} \cmidrule(l){8-8} 
     ResNet50 & \bf 72.32 / 61.82 & 75.26 / 67.21 & \bf 88.96 / 87.61 & 77.20 / 70.52 & 61.62 / 37.88 & 58.35 / 28.82 & 72.28 / 58.98\\
         Xception &  52.05 / 07.98 &   58.60 / 29.41  & 54.62 / 16.99 & 60.01 / 33.43 & 71.53 / 60.03 & 51.64 / 06.66 & 58.08 / 25.75\\
        DeiT     &  50.40 / 02.01 &   50.18 / 01.17  & 50.14 / 01.01 & 56.25 / 22.54 & 93.26 / 92.79 & 79.84 / 74.82 & 63.34 / 32.39\\ \midrule
FIDAVL & 67.35 / 56.01 & \bf 86.56 / 85.61 & 81.38 / 78.91 & \bf 94.93 / 95.02 & \bf 96.25/ 96.36 &  \bf 89.78 / 88.98 & \bf 86.04 / 83.48 \\ \bottomrule
\end{tabular}
\end{adjustbox}
\begin{flushleft}
    \scriptsize $^\star$ Diffusion-based model.  $^\oplus$ GAN-based model.
\end{flushleft}
\end{table*}

Results in Table~\ref{tab:detectionTaskGeneralization} highlight \acs{fidavl}'s generalization performance across the different subsets. Overall, \acs{fidavl} generalizes very well, with an average accuracy of 86.04\% and F1-score of 83.48\% across all unseen test sets during training. 

ResNet50 demonstrates moderate performance across subsets, showing notable strength in ADM and IDDPM, while Xception exhibits variable performance, particularly struggling with ADM, DDPM, and IDDPM subsets. DeiT performs similarly to Xception, facing challenges in ADM, DDPM, and IDDPM subsets. \acs{fidavl} shows superior performance across most subsets, especially excelling in DDPM, IDDPM, PNDM, and GAN-based subsets like Diff-StyleGAN2 and ProjectedGAN.

Moreover, the results reveal patterns and considerations that need further investigation:

\begin{itemize}
\item  ADM$^\star$ subset: \acs{fidavl} achieves an accuracy of 67.35\% and F1-score of 56.01\%, indicating moderate performance.
\item  DDPM$^\star$ subset: \ac{fidavl} achieved a commendable accuracy of 86.56\% and an F1-score of 85.61\%, suggesting strong performance in detecting diffusion-based models. However, deeper analysis is warranted to understand any potential biases or limitations when handling these types of synthetic images.
\item IDDPM$^\star$ subset: \ac{fidavl}'s performance (accuracy: 81.38\%, F1-score: 78.91\%) indicates slightly reduced effectiveness compared to other subsets, suggesting potential challenges in detecting specific characteristics associated with this subset, and necessitating further investigation into the model's adaptability.
\item PNDM$^\star$ subset: \ac{fidavl} excelled with an impressive accuracy of 94.93\% and an F1-score of 95.02\%, indicating robust performance in detecting certain types of diffusion-based models. Besides, this highlights its strengths but raises questions about its generalizability across all diffusion-based variants.
\item Diff-StyleGAN2$^\oplus$ subset: \ac{fidavl} demonstrated high accuracy (96.25\%) and a high F1-score (96.36\%) in detecting this \ac{gan}-based model. Although this achievement underlines the ability of \ac{fidavl} to identify this specific \ac{gan} architecture, further research is needed to assess its performance over a wider range of \ac{gan} variations.
\item ProjectedGAN$^\oplus$ subset: \acs{fidavl} demonstrates strong performance with an accuracy of 96.38\% and an f1-score of 96.49\%. This showcases \acs{fidavl}'s ability to accurately detect images generated by ProjectedGAN models.
\end{itemize}

Although \ac{fidavl} shows promising performance, a rather critical aspect deserves closer investigation. \ac{fidavl}'s exceptional performance on certain subsets raises questions about its focus on specific model characteristics versus broader synthetic image detection. However, the balance between model specificity and general applicability is essential for its deployment in the real world. The results underline \acs{fidavl}'s effectiveness in handling diverse synthetic image datasets generated by unseen models. Its superior performance signifies strong generalization potential, critical for real-world applications where model adaptability to varying synthetic data sources is essential.

\subsection{Synthetic Image Attribution}
\vspace{-1.0mm}
In this section, we assess the performance of \acs{fidavl} in the synthetic image attribution task using ROUGE scores as metrics, in conjunction with standard classification metrics such as accuracy and F1-score. As detailed in Subsection~\ref{sec:problemFormulation}, \acs{fidavl} generates text as output. ROUGE scores are widely recognized as metrics commonly used in text generation tasks. These scores primarily gauge the quality of machine-generated text by comparing it to reference text, measuring various aspects of text similarity, such as overlap in n-grams (consecutive sequences of words). Furthermore, the inclusion of accuracy and F1-score provides a comprehensive understanding of \acs{fidavl}'s performance in synthetic image attribution. In our experiment, we utilize two ROUGE scores: ROUGE-2 and ROUGE-L.

\begin{table*}[]
\caption{Performance evaluation of synthetic image attribution task.}

\label{tab:attributionTask}
\begin{adjustbox}{max width=\linewidth}
\begin{tabular}{@{}c|c|c|c|c|c|c|c@{}}
\toprule
\multirow{2}{*}{Method} &
  \multicolumn{6}{c}{ROUGE-2 / ROUGE-L scores on different testing subsets} &
  \multirow{2}{*}{\begin{tabular}[c]{@{}c@{}}Average\\ (in \%)\end{tabular}} \\ \cmidrule(lr){2-7}
 &
  LDM$^\star$ &
  SD v1.4$^\star$ &
  GLIDE$^\star$ &
  ProGAN$^\oplus$ &
  StyleGAN$^\oplus$ &
  Diff-ProjectedGAN$^\oplus$ &
   \\ \cmidrule(r){1-1} \cmidrule(l){8-8} 
FIDAVL   & 92.23 / 94.82 & 97.39 / 98.19 & \bf 97.41 / 98.20  &  94.99 / 97.01 & 93.21 / 96.14 & 90.62 / 94.64 & 94.30 / 96.50 \\
\end{tabular}
\end{adjustbox}
\begin{adjustbox}{max width=\linewidth}
\begin{tabular}{@{}c|c|c|c|c|c|c|c@{}}
\toprule
\multirow{2}{*}{Method} &
  \multicolumn{6}{c}{ACC / F1-score on different testing subsets} &
  \multirow{2}{*}{\begin{tabular}[c]{@{}c@{}}Average\\ (in \%)\end{tabular}} \\ \cmidrule(lr){2-7}
 &
  LDM$^\star$ &
  SD v1.4$^\star$ &
  GLIDE$^\star$ &
  ProGAN$^\oplus$ &
  StyleGAN$^\oplus$ &
  Diff-ProjectedGAN$^\oplus$ &
   \\ \cmidrule(r){1-1} \cmidrule(l){8-8} 
FIDAVL   & 87.89 / 89.27 & 96.10 / 97.96 & \bf 96.12 / 98.00  &  87.39 / 93.17 & 84.57 / 90.95 & 77.92 / 86.54 & 88.33 / 92.64\\ \bottomrule
\vspace{-5mm}
\end{tabular}
\end{adjustbox}
\begin{flushleft}
    \scriptsize $^\star$ Diffusion-based model.  $^\oplus$ GAN-based model.
\end{flushleft}
\end{table*}

Table~\ref{tab:attributionTask} presents a comprehensive evaluation of \acs{fidavl} in synthetic image attribution task across different test sets classified according to their underlying architectures: diffusion models (LDM, Stable Diffusion v1.4, GLIDE) and \acs{gan} models (ProGAN, StyleGAN, Diff-ProjectedGAN). The evaluation metrics used are ROUGE-2, ROUGE-L, accuracy, and F1-score, measured on different test subsets.

First, the results show that \acs{fidavl} generally achieves competitive performance in terms of ROUGE scores, accuracy, and F1-score on diffusion-based models compared to \acs{gan}-based models. In particular, Stable Diffusion v1.4 and GLIDE achieve higher ROUGE scores, accuracy and F1-score than ProGAN, StyleGAN, and Diff-ProjectedGAN. This variation highlights the sensitivity of \acs{fidavl} to the characteristics inherent in different architectural models, potentially indicating the model's proficiency in specific image generation paradigms.

\begin{figure}[!ht]
    \centering
    \includegraphics[width=0.82\linewidth]{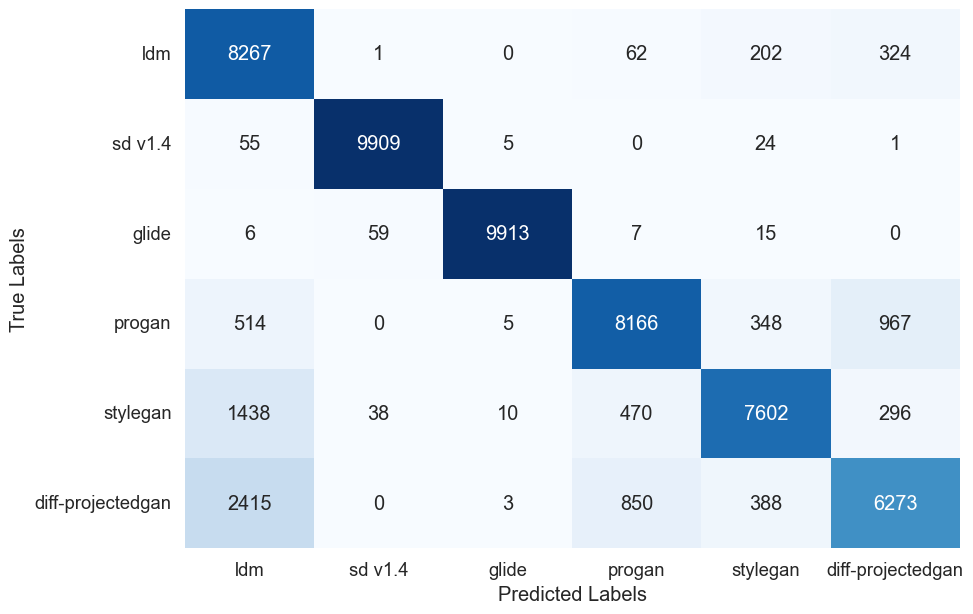}
    \caption{Confusion Matrix for Attribution Task: Synthetic data correctly classified as synthetic but attributed to a different source from the generating source.}
    
    \label{fig:confmatricAttribution}
\end{figure}

Fig.~\ref{fig:confmatricAttribution} illustrates the distribution of accurately classified synthetic images across various generative models. The diagonal elements (True Positive) depict the number of correct predictions for each category. Remarkably, \acs{fidavl} demonstrates exceptional performance on stable diffusion and Glide, with $9909$ and $9913$ instances correctly classified, respectively. 
However, the matrix also sheds light on areas of concern. \acs{fidavl} encounters difficulties in accurately attributing \acs{gan}-based generated images to their specific source models. Many \acs{gan}-based generated images are incorrectly attributed to \acs{ldm} and other \acs{gan}-based models. This may be attributed to the fact that unconditional diffusion models, such as \acs{ldm}, share similarities with \acs{gan}-based generative models, posing challenges for accurate attribution.

\vspace{-6mm}
\section{Conclusion and Future Work} 
\label{sec:conclusion}
\vspace{-6.0mm}
In this paper, we have proposed \acs{fidavl}, a novel multitask framework for \ac{ai}-generated image detection and attribution leveraging vision-language models. Through the integration of vision and language modalities, \acs{fidavl} exhibited exceptional performance in accurately discerning and attributing \ac{ai}-generated images to their respective source models. Extensive experimentation validated the effectiveness of \acs{fidavl} in addressing the challenges of synthetic image detection and attribution simultaneously. Our findings underlined the significance of interdisciplinary approaches in tackling complex problems in today's rapidly evolving technological landscape. With its promising performance, \acs{fidavl} presented a valuable solution to enhance accountability and trust amidst the proliferation of fake images.
In future endeavors, we aim to conduct additional experiments to evaluate the robustness and generalization capabilities of \acs{fidavl} in real-world scenarios. This includes exploring scenarios involving JPEG compression, scaling, unseen images from new generative models, and added noise. Additionally, we plan to extend \acs{fidavl} into a multi-head vision-language framework to further enhance its capabilities and versatility.

\noindent {\bf Acknowledgments:} This work has been partially funded by the project PCI2022-134990-2 (MARTINI) of the CHISTERA IV Cofund 2021 program. Abdenour Hadid is funded by TotalEnergies collaboration agreement with Sorbonne University Abu Dhabi.


\bibliographystyle{splncs04}
\bibliography{ref}

\end{document}